\documentclass[final]{cvpr}

\usepackage{times}
\usepackage{epsfig}
\usepackage{graphicx}
\usepackage{amsmath}
\usepackage{amssymb}
\usepackage{booktabs}

\usepackage{dsfont}

\usepackage{diagbox}
\usepackage{multirow}
\usepackage{xcolor}
\usepackage{algorithm}
\usepackage{algorithmic}

\newcommand{\R}{\mathbb{R}}
\newcommand{\bb}{\boldsymbol{b}}
\newcommand{\bff}{\boldsymbol{f}}
\newcommand{\bc}{\boldsymbol{c}}
\newcommand{\br}{\boldsymbol{r}}

\newcommand{\bq}{\boldsymbol{q}}

\newcommand{\Q}{\mathcal{Q}}

\newcommand{\N}{\mathcal{N}}

\usepackage[pagebackref=true,breaklinks=true,colorlinks,bookmarks=false]{hyperref}

\def\cvprPaperID{****} 
\def\confName{ICCV}
\def\confYear{2021}
\graphicspath{{./figs/}{./figs/result/}}

\begin{document}

\title{Graph-DETR3D: Rethinking Overlapping Regions \\for Multi-View 3D Object Detection}

\author{Zehui Chen$^1$ \quad Zhenyu Li$^{2}$ \quad Shiquan Zhang$^{3}$ \quad Liangji Fang$^{3}$ \quad Qinhong Jiang$^{3}$ \quad Feng Zhao$^1$ \\
$^{1}$University of Science and Technology of China \\ $^{2}$Harbin Institute of Technology  \quad $^{3}$SenseTime Research}

\maketitle

%

\begin{abstract}
    3D object detection from multiple image views is a fundamental and challenging task for visual scene understanding. Due to its low cost and high efficiency, multi-view 3D object detection has demonstrated promising application prospects. However, accurately detecting objects through perspective views in the 3D space is extremely difficult due to the lack of depth information. 
Recently, DETR3D \cite{wang2022detr3d} introduces a novel 3D-2D query paradigm in aggregating multi-view images for 3D object detection and achieves state-of-the-art performance. In this paper, with intensive pilot experiments, we quantify the objects located at different regions and find that the ``truncated instances'' (\textit{i.e.,} at the border regions of each image) are the main bottleneck hindering the performance of DETR3D. Although it merges multiple features from two adjacent views in the overlapping regions, DETR3D still suffers from insufficient feature aggregation, thus missing the chance to fully boost the detection performance. In an effort to tackle the problem, we propose Graph-DETR3D to automatically aggregate multi-view imagery information through graph structure learning (GSL). It constructs a dynamic 3D graph between each object query and 2D feature maps to enhance the object representations, especially at the border regions. Besides, Graph-DETR3D benefits from a novel depth-invariant multi-scale training strategy, which maintains the visual depth consistency by simultaneously scaling the image size and the object depth. Extensive experiments on the nuScenes dataset demonstrate the effectiveness and efficiency of Graph-DETR3D. Notably, our best model achieves 49.5 NDS on the nuScenes test leaderboard, achieving new state-of-the-art in comparison with various published image-view 3D object detectors.
\end{abstract}

\section{Introduction}

\begin{figure}[!t]
	\centering
    \includegraphics[width=1.0\columnwidth]{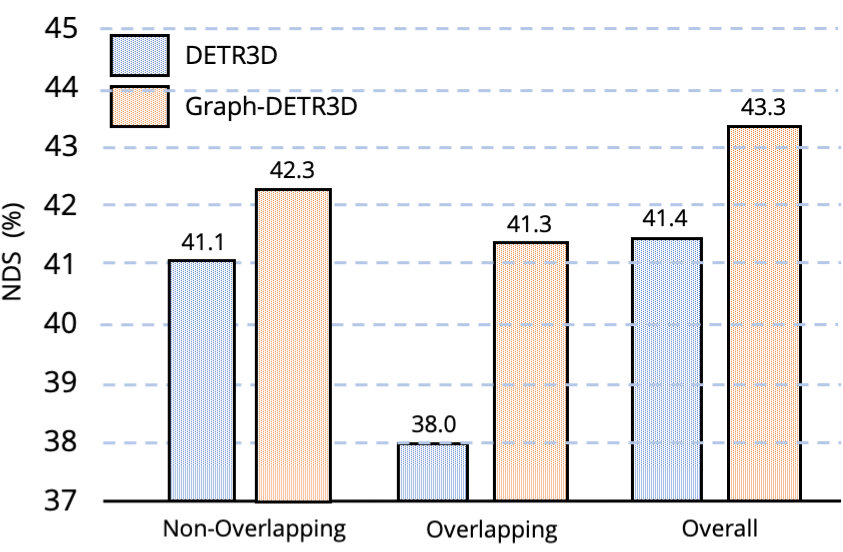}
    \caption{Comparison between Graph-DETR3D and its baseline on nuScenes validation subset. Graph-DETR3D boosts the detection accuracy on overlapping regions by a remarkable margin.}
    \label{fig:intro}
 \end{figure}
  
3D object detection, localizing objects in the 3D space, is an indispensable ingredient for 3D scenes understanding and widely adopted in various real-world applications, such as augmented reality (AR) \cite{azuma1997survey,carmigniani2011augmented,azuma2001recent} and autonomous driving \cite{chen2022autoalign,chen2017multi,qi2018frustum}. Though promising results have been achieved by utilizing LiDAR devices \cite{zhou2018voxelnet,yin2021center,wang2021object,pan20213d}, these methods suffer from the high cost of deploying devices and the sparsity of input LiDAR points. Alternatively, cameras are economical, easy-to-deploy, and light-weight choices for 3D object detection. Recently, multi-view 3D object detection, aiming at detecting objects though multiple surrounding images, has received increasing attentions due to its simplicity and elegant pipeline \cite{wang2022detr3d,huang2021bevdet}. However, the accuracy of such approaches is far lagged behind, which hinders them from practical applications. 

Previous state-of-the-art methods  mainly attribute the poor detection accuracy to the localization error due to the ill-posed depth estimation from monocular images. Hence, they either introduce an external depth estimation network \cite{reading2021categorical,you2019pseudo,wang2021depth} or incorporate prior visual knowledge \cite{lu2021geometry,li2019gs3d}. Recently, by extending from DETR~\cite{carion2020end}, DETR3D \cite{wang2022detr3d} provides a simple and effective framework for multi-view 3D object detection. It introduces a novel concept of \textit{reference point}, which is generated from each object query, to iteratively aggregate 2D features. By projecting these points into the image plane with the camera projection matrices, the object query is able to perceive imagery views and therefore localize the objects accurately. In this work, we carefully quantify the objects located at different regions in DETR3D and find that the ``truncated instances'' account for the main localization errors. Although DETR3D utilizes the reference point to naturally bridge the features from multiple images, we notice that it still suffers from insufficient information aggregation: only the center coordinate of the prediction is used as the sampling point to update the representation of the object query. This implementation sounds reasonable but will suffer from the inevitable dilemma. Think of such a phenomenon: when the centroid of the truncated object is only projected into one image plane, it loses the natural benefit enjoyed by multi-view detection (\textit{i.e.,} to perceive information in the overlapping regions from different perspectives). Besides, these truncated objects are usually close to the ego point, posing potential safety problems if not well localized. Thus, how to fully leverage the overlapping regions to boost the detection accuracy on truncated objects for multi-view 3D object detection remains a vital problem. 

In this paper, we propose a learnable approach, namely \textbf{Graph-DETR3D}, based on graph structure learning (GSL), to dynamically aggregate multiple imagery information for 3D object detection (Figure \ref{fig:intro}). Different from the vanilla DETR3D, which establishes a one-to-one (each object query only corresponds to a single reference point) matching diagram, we treat this process as a graph message propagation. Specifically, we construct the edge connections between each object query and the input 2D feature maps through a dynamic graph lying in the 3D space. For each object query, we first compute its reference point and dynamically decide the $k$ neighbors in the spatial space to build a 3D graph, where each graph node is projected into the camera view for image feature sampling. The sampled features are then merged together through graph affinity matrices, to further enhance the representation of object queries. Such a paradigm brings in two benefits: (i) the 3D graph structure enables the object query to perceive sufficient imagery information through its graph nodes and implicitly enhances the visual consistency from multi-camera views, especially for overlapping regions, and (ii) the dynamic graph propagation aggregates neighbor contexts efficiently and adaptively by adjusting the affinity matrices automatically. Moreover, data augmentation, especially multi-scale training~\cite{huang20201st}, is a crucial step for current detectors to achieve competitive results. However, directly applying multi-scale training on vision-based 3D detectors can severely deteriorate the model performance. After careful analysis, we find a simple correlation between the image scale and the object depth by restricting the scaling pattern. Based on this finding, we introduce a novel depth-invariant multi-scale training strategy to further boost the detection accuracy. 

 In summary, the main contributions of this work are four-fold: (1) By quantifying the objects located in different regions, we identify the `truncated instances' accounting for the main localization error in vision-based 3D detection approaches. (2) We propose a novel graph-based multi-view 3D object detection framework, namely Graph-DETR3D. It builds the connections between the object query and the surrounding image views with a dynamic graph, through which the model is able to fully leverage 2D information, especially in the overlapping regions. (3) We introduce a simple yet effective depth-invariant multi-scale training strategy for vision-based 3D detectors, to overcome the depth ambiguity issue when resizing the images. (4) Through extensive experiments, we validate the effectiveness and generality of the proposed Graph-DETR3D. Notably, our best model achieves new state-of-the-art performance on the competitive nuScenes dataset.
    
\section{Related Work}

\begin{figure*}[!t]
    \includegraphics[width=2.1\columnwidth]{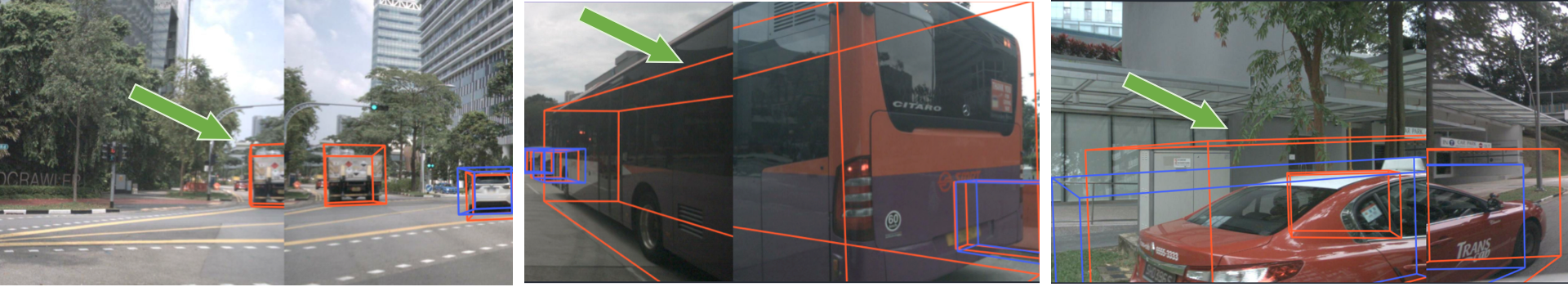}
    \caption{Visualization of \textcolor[RGB]{0,179,0}{failure cases} with DETR3D on truncated objects. Each image is cropped from two adjacent camera images. Best view in color: the prediction and ground truth are in \textcolor{blue}{blue} and \textcolor{orange}{orange}, respectively.}
    \label{fig:error}
 \end{figure*}
 
%
\subsection{Monocular 3D object detection} Monocular detection has been extensively studied since it was first proposed in 2016 \cite{chen2016monocular}. Previous works \cite{qin2019monogrnet,shi2021geometry,simonelli2019disentangling} mainly follow the paradigm of localizing the 2D bounding boxes and then predicting the 3D information through RoI features. MonoGRNet \cite{qin2019monogrnet} formulates such a task into a 9 degree-of-freedom optimization problem and leverages geometrical constraints to improve the depth prediction accuracy. FQNet \cite{liu2019deep} explicitly estimates the prediction quality measured by 3D IoU score to enhance the final performance. Several recent works explore the single-stage architecture for accurate 3D detection. FCOS3D \cite{wang2021fcos3d} constructs a fully-connected monocular 3D detection framework by extending FCOS from 2D detection. PGD \cite{wang2022probabilistic} incorporates the concept of uncertainty of depth estimation and further advances the detection accuracy. In \cite{ma2021delving}, the localization error is proven to be the most vital problem. Therefore, MonoDLE proposes to disentangle the training samples located at different regions. An alternative solution is to explicitly (implicitly) utilize depth information. Pseudo-LiDAR \cite{wang2019pseudo} reconstructs the point cloud by estimating the pixel-wise depth and then applies LiDAR-based approaches. Instead of directly transforming pixels into the 3D space, $D^4LCN$ \cite{ding2020learning} generates depth-guided convolutional kernels to extract the imagery information for 3D object detection.
\subsection{Multi-View 3D object detection} Actually, multi-view detection can also be solved by applying monocular 3D detectors on each single view and then ensembling them together \cite{park2021pseudo,wang2021fcos3d}. However, such a paradigm is a sequential process and cannot fully leverage the benefits provided by multi-view images. Following DETR \cite{carion2020end}, DETR3D \cite{wang2022detr3d} presents a novel query-based detection framework. It constructs relationships between object query and reference point, which are iteratively projected back to the camera views and sample the 2D features. BEVDet \cite{huang2021bevdet} employs the Lift-Splat-Shoot \cite{philion2020lift} to transform imagery information to the BEV space and directly outputs the predictions with a CenterPoint \cite{yin2021center} head.
 
\subsection{Graph Structure Learning}
The graph structure learning (GSL) targets at jointly learning an optimized graph structure and its corresponding representations. Most existing GSL models \cite{zhao2021heterogeneous,luo2020parameterized,zhu2021deep} follow a three-stage pipeline: (i) graph construction, (ii) graph structure modeling, and (iii) message propagation. To build an initial graph, previous works use either k-nearest neighbors \cite{preparata2012computational} or $\epsilon$ proximity thresholding \cite{bentley1977complexity}. After that, a structure learner $g$ is designed to refine the preliminary graph through a metric function \cite{park2020agcn,wang2020gcn} or learnable parameters \cite{kreuzer2021rethinking,sun2021graph}. Once obtaining an optimized adjacency matrix provided by $g$, it employs a GNN encoder $f$ to propagate the node features to its refined neighborhoods. It is common to iteratively repeat the last two steps to reach the optima of the final graph topologies and representations.
\section{Error Analysis}

In this section, we explore the key factor that hinders the performance of current vision-based 3D detectors on localization error. Inspired by \cite{ma2021delving}, we conduct an error analysis on the objects located at different regions on nuScenes validation subset. Different from the previous experiments, which only evaluate the objects based on their distances to the ego point, we split them with respect to the camera viewpoint. Specifically, we categorize the objects depending on the number of images its centroid is projected into, and evaluate the performance on them. An illustration of the region split is depicted in the appendix. Normally, we find that if any of the 8 corners of one instance is projected into two adjacent images (\textit{i.e.,} the overlapping region), this object is more prone to be truncated, since it mainly locates at the side-border region of the image. We select two representative vision-based 3D detectors, FCOS3D (monocular based) \cite{wang2021fcos3d} and DETR3D (multi-view based) \cite{wang2022detr3d} and report the results in Table \ref{tab:nus_pilot}. From the table, we notice that objects located in the non-overlapping regions are highly consistent with the overall performance. However, both approaches fail to handle instances in the overlapping regions: they all incur significant accuracy decline (roughly 9.0 mAP and 5.0 NDS). Such a relative 10\% decline implies the inferior ability of current models on localizing truncated objects. To better illustrate such a phenomenon, we also visualize several failure cases predicted by the competitive DETR3D baseline in Figure \ref{fig:error}. Compared with the ill-posed depth estimation from monocular images, it is more feasible to improve the detection accuracy in the overlapping region.

\begin{table}
	\centering
		\setlength{\tabcolsep}{0.5mm}
			\begin{tabular}{c | c c | c c}
				\toprule
				\multirow{2}*{Region}  &  \multicolumn{2}{c|}{FCOS3D} & \multicolumn{2}{c}{DETR3D}\\
				    & mAP (\%) & NDS (\%) & mAP (\%)  & NDS (\%) \\
				\midrule
				Overall & 32.1 & 39.5 & 32.5 & 39.8\\
				Non-Overlap & 32.2 \footnotesize{(\textcolor{cyan}{$+0.1$})} & 39.5 \footnotesize{(\textcolor{cyan}{$+0.0$})} & 32.9 \footnotesize{(\textcolor{cyan}{$+0.4$})} & 39.9 \footnotesize{(\textcolor{cyan}{$+0.1$})}\\
				Overlap & 22.6 \footnotesize{(\textcolor{orange}{$-9.5$})} & 33.6 \footnotesize{(\textcolor{orange}{$-5.9$})} & 23.3 \footnotesize{(\textcolor{orange}{$-9.2$})} & 35.0 \footnotesize{(\textcolor{orange}{$-4.8$})}\\
				\bottomrule
			\end{tabular}
	\vspace{0.2pt}
	\caption{Detection accuracy in different regions with two different vision-based 3D detectors, FCOS3D and DETR3D, on nuScenes validation subset. }
	\label{tab:nus_pilot}
	\vspace{-0.5pt}
\end{table}
\section{Methodologies}

\begin{figure*}[!t]
	\centering
    \includegraphics[width=0.9\linewidth]{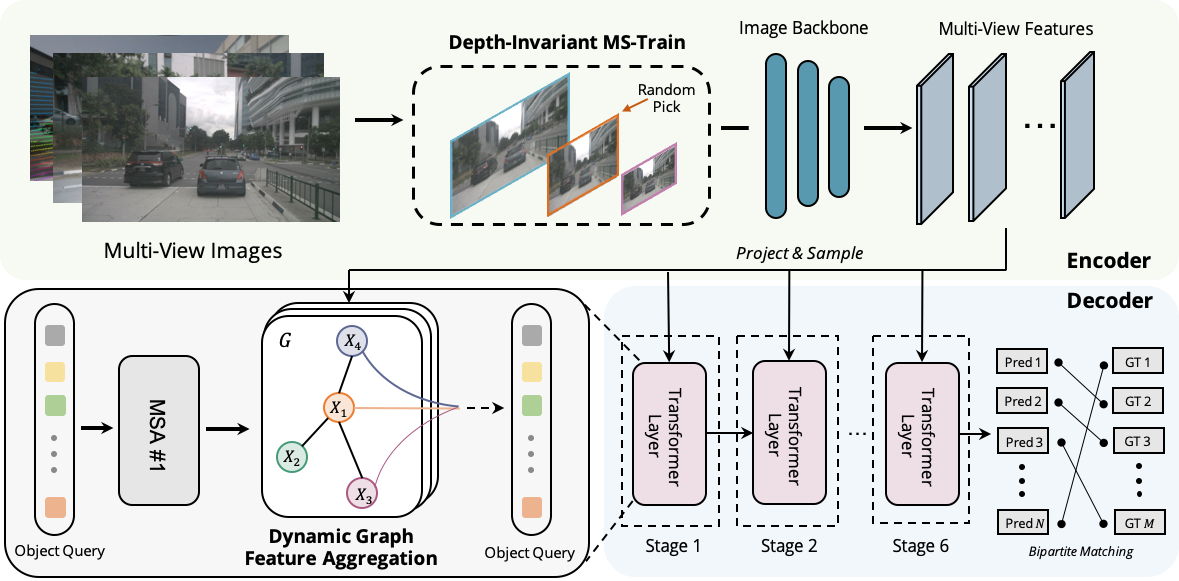}
    \caption{The framework overview of Graph-DETR3D. It differs from its vanilla baseline in two aspects: (1) dynamic graph feature aggregation module (Section \ref{sec:dgfa}), and (2) depth-invariant multi-scale training strategy (Section \ref{sec:mstrain}).}
    \label{fig:framework}
 \end{figure*}
 
\subsection{Preliminaries}

\noindent\textbf{DETR3D.} It first extracts 2D features $\mathbf{F}$ from multiple surrounding images with a ResNet backbone and a FPN. Then, it adopts a set of object queries $\bq$, linking with 3D reference points, to index into 2D features $\mathbf{F}$ through camera projection matrices. Such a paradigm enables each query to iteratively aggregate relevant 2D information with bilinear interpolation. Besides, the queries also sequentially interact with each other through a transformer-based self-attention operation. Finally, DETR3D yields the object class as well as its localization in 3D space from each query, supervised by a set-to-set bipartite loss. 

\noindent\textbf{Graph Structure Learning.} Let $\mathcal{G} = (\bf{A}, \bf{X}, \bf{V})$ denote a graph, where $\textbf{A} \in \R^{N\times N}$ is the adjacency matrix, $\textbf{V} =\{v_i\}_1^N$ is the $N$ number of nodes in the graph, $\textbf{X} \in \R^{N\times F}$ is the node feature matrix with the $i$-th entry $x_i \in \R^F$ representing the attribute of node $v_i$. Given the graph $\mathcal{G}$ at the beginning, graph structure learning is to simultaneously learn the adjacency matrix $A^{\star}$ and its corresponding graph representation $X^{\star}$. Specifically, $A$ is a binary or learnable matrix to describe the connection between graph nodes and the model aims to optimize the representation of each graph node $x_i$ by extracting structured information through the whole propagation network. Given the feature vector $x_i$ of node $v_i$, it samples $K$ one-hop neighbor nodes $\N(i)$ defined by the adjacency matrix $A$ and then compute the message propagation $G$ by,
\begin{align}
	x_i^{new} & = x_i + G(\textbf{A}\{v_i; \N(i)\}, \textbf{W})\\
	 & = x_i + \sum_{j\in \N(i)}A_{i,j} \cdot x_j \cdot w_{i,j},
\end{align}
where $A_{i,j}$ and $w_{i,j}$ are the connection intensity and transformation matrix between the node $v_i$ and $v_j$.

\subsection{Framework Overview}
An overview of the proposed Graph-DETR3D is depicted in Figure \ref{fig:framework}. It follows the basic structure of DETR3D, consisting of three components: an image encoder, a transformer-style decoder, and object prediction heads. Given with a set of images $\mathbf{I} = \{I_1, I_2, ..., I_K\} \in \R^{H\times W \times 3}$ (captured by $N$ surrounding cameras), Graph-DETR3D aims to predict the localization and categories of the interested bounding boxes. These images are first processed with an image encoder, including a ResNet \cite{he2016deep} and a FPN \cite{lin2017feature}, into a set of features $\textbf{F}$ with respect to $L$ feature map levels. Then, we construct a dynamic 3D graph to extensively aggregate 2D information through our proposed dynamic graph feature aggregation module to refine the representation of object queries. Finally, the enhanced object queries are utilized to output the final prediction.

\subsection{Dynamic Graph Feature Aggregation}
\label{sec:dgfa}

\subsubsection{Revisiting to DETR3D-style feature aggregation}
We first revisit the multi-view feature aggregation proposed in DETR3D. It constructs a concept of reference point linked with the object query. By projecting the reference point back into the image space with the camera projection parameters, DETR3D samples the 2D features from all camera views. Specifically, it first initializes a set of object queries $\Q=\{\bq_{1}, \ldots, \bq_{M^*}\}\subset\R^{C}$ similar to \cite{wang2022detr3d}. Then, it decodes the reference point $\bc_i$ from each object query $\bq_i$, which can be viewed as the center of the $i$-th predicted box:
\begin{equation}
	\bc_i = \Theta^{ref}(\bq_i),
\end{equation} 
where $\Theta^{ref}$ is a neural network. Next, $\bc_i$ is projected into each $N$ images using the camera transformation matrix $T_n$ and the $l$-th level image features $\textbf{F}_{nl}$ from each camera view $n$ are collected with bilinear interpolation:
\begin{align}
	\br_{ni} & = T_n \cdot c_i,\\
	\bff_{n i l}  & = f^{\mathrm{bilinear}}(\textbf{F}_{nl}, \br_{n i}),
\end{align}
where $\br_{in}$ is the 2D coordinate of the reference point $\br_{i}$ at the $n$-th camera. After that, DETR3D merges multi-view features to generate the new object query in a residual manner:
\begin{align}
	\bff_i = \frac{1}{\sum_L \sum_N \sigma_{n i l}} \sum_L \sum_N	\bff_{n i l} \sigma_{n i l},\\
	\bq_i^{new} = \bq_i + \bff_i, \hspace{50pt}
\end{align}
where $\sigma_{i l n}$ is a binary value to indicate whether a reference point is projected inside an image plane. 

\subsubsection{Dynamic Graph Feature Aggregation}
 
\begin{figure}[!t]
    \includegraphics[width=\columnwidth]{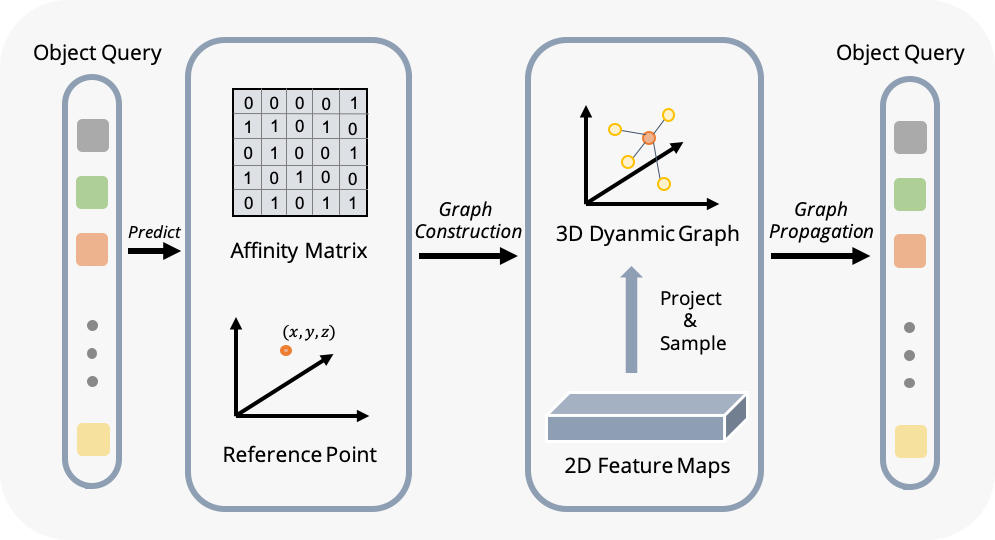}
    \caption{Illustration of the dynamic graph feature aggregation module. It first constructs 3D graphs for each reference point, and then aggregates relevant imagery information through graph connections to enhance the representation of the object query.}
    \label{fig:dgfa}
 \end{figure}

The bottleneck of DETR-3D is that it fails to consider context features for sufficient geometrical and semantical relevant information. In this part, we propose a novel dynamic graph feature aggregation (DGFA) module, which automates the multi-view feature fusion process through a learnable 3D graph. Figure \ref{fig:dgfa} shows the dynamic graph feature aggregation process. We first construct a learnable 3D graph for each object query, and then sample features from 2D image planes. Finally, the representation of object query is enhanced through graph connections. Such a mutually connected message propagation scheme enables an iterative refinement scheme on graph structure construction and feature enhancement.

More formally, given the reference point $c_i = (x, y, z)$ predicted from the network, we regard its spatial neighbors as possible graph nodes and connect them via adjacency matrix $A$ predicted by the network. Note that the created graph is asymmetric, \textit{i.e.,} an edge from node $a$ to $b$ does not imply the existence of an edge from $b$ to $a$. To dynamically sample context-relevant nodes for each object query, we enable the network to walk around the uniform space so as to determine the most informative neighbors for feature aggregation. Suppose $\Delta_i \in \R^{K\times 3}$ is the predicted walk for connected graph nodes, the set of connected graph nodes $\N(i)$ of each query $q_i$ can be described as:
\begin{align}
	\N(i) = c_i + \Delta_i = c_i + f(q_i),
\end{align}
where $f$ is an MLP to generate the transformation offset for the sampling nodes based on each object query $q_i$. Considering that only the reference point features will be collected by each object query, we omit the edge connections between two nodes if both are not reference point for computational efficiency. 

After constructing the graph, we adopt a CNN structure as the unary model to enhance the features for each query. By projecting the neighbors to 2D image plane for feature sampling, we can obtain the hidden state $x_j$ of each graph node $v_j$:
\begin{align}
	r_{jn} = T_n \cdot v_j,~~~ v_j \in \N(i) \hspace{50pt}\\
	x_j = \frac{1}{\sum_1^L \sum_1^N \sigma_{jnl}} \sum_1^N \sum_1^L f^{\text{bilinear}}(\textbf{F}_{ln}, r_{jnl}) \sigma_{jnl}.
\end{align}
Given that the 3D graph already contains the spatial encodings, this graph connection explores not only the appearance but also the geometric information. The final propagation process at iteration $t+1$ can be expressed as
\begin{align}
	q_i^{t+1} & = q_i^t + G(\textbf{A}\{c_i; \N(i)\}, \textbf{W}),\\
		& = q_i^t + \sum_{j\in \N(i)} x_j \cdot w_{i,j},
\end{align}
where $w_{i,j}$ is the edge weights between the object query $q_i$ and one-hop neighbor node $v_j$ predicted from $q_i$ with an MLP layer.

\subsection{Depth-Invariant Multi-Scale Training}
\label{sec:mstrain}

\begin{figure}[!t]
    \includegraphics[width=1.0\columnwidth]{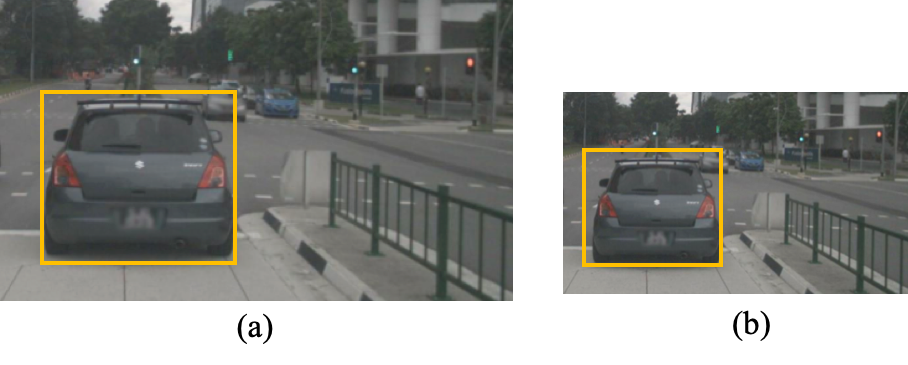}
    \vskip -1em
    \caption{Visualization of the scale ambiguity issue in multi-scale training. Though the number of image pixels inside the yellow box in (a) and (b) varies, the supervision remains the same.}
    \label{fig:scale_issue}
    \vskip -1em
 \end{figure}

Multi-Scale training \cite{singh2018sniper} is a commonly adopted data augmentation strategy in 2D and 3D object detection tasks, which is proven to be effective and inference cost-free. However, it hardly appears in vision-based 3D detection approaches. Considering that diverse input image sizes can improve the model robustness, we implement the vanilla multi-scale training strategy, by simultaneously rescaling the image sizes and modifying the camera intrinsics. An interesting phenomenon is that the final performance incurs a drastic decline. By carefully analyzing the input data, we find that simply rescaling the image leads to an perspective-ambiguous issue: when the object is resized to a larger/smaller scale, its absolute attributes (\textit{i.e.}, size of the object, distance to the ego point) are not changed. As a concrete example, we show this ambiguous problem in Figure \ref{fig:scale_issue}. Despite the absolute 3D position of the selected regions in (a) and (b) are the same, the number of image pixels varies. As analyzed in \cite{dijk2019neural}, the depth prediction network tends to estimate the depth based on its occupied area in the image. Therefore, such a training pattern in Figure \ref{fig:scale_issue} can confuse the model on depth prediction and further deteriorate the final performance.

To overcome the problem, we recompute the ground truth depth from the perspective of image pixels. Specifically, instead of viewing the depth prediction problem as one absolute value, we first decode it into a pixel-level depth, similar in \cite{park2021pseudo}:
\begin{align}
	d &= \frac{1}{p} \cdot (\sigma_l \cdot z + \mu_l),\\
	p &= \sqrt{\frac{1}{f_x^2} + \frac{1}{f_y^2}},
\end{align}
where $z$ is the expected depth prediction, $d$ is the metric depth required in object localization. $(\sigma_l, \mu_l)$ are learnable scalers parameterized by a simple neural layer and $p$ is the pixel size computed from the focal lengths, $f_x$ and $f_y$ are the constant value. Suppose the $x$ side is resized by a factor of $r_x$ and $y$ side by $r_y$, then the recomputed pixel-size can be formulated as:
\begin{align}
	p_{new} = \sqrt{\frac{1}{(r_x f_x)^2} + \frac{1}{(r_y f_y)^2}}.
\end{align}
To avoid shape deformation of the object, we stipulate $r = r_x = r_y$, therefore, $p_{new}$ and the final expected depth prediction $z_{new}$ can be simplified to: 
\begin{align} 
	p_{new} = \frac{1}{r} \sqrt{\frac{1}{f_x^2} + \frac{1}{f_y^2}} = \frac{1}{r} p\\
	z_{new} = z / r.\hspace{30pt}
\end{align}

To this end, we propose the depth-invariant multi-scale training strategy, by simultaneously rescaling the image size and the object depth. The detailed implementation is shown in Algorithm \ref{algo:mstrain}.

\begin{algorithm}[!t]
\caption{Depth-Invariant Multi-Scale Training}
\begin{algorithmic}[1]
\REQUIRE Multi-View Images $\mathbf{I} = \{I_1, I_2, ..., I_N\}$, Ground Truth 3D Object Center Set $\mathbf{O} = \{O_1, O_2, ..., O_M\}$, Rescale Range [$r_{min}, r_{max}$]. 
\STATE $r$ $\leftarrow$ SAMPLE($r_{min}, r_{max}$);
\FORALL{$n$ such that $I_n \in \mathbf{I}$}
    \STATE $//$ resize the image
    \STATE $I_n \leftarrow RESIZE(I_n, \sigma)$;
\ENDFOR
\FORALL{$i$ such that $O_m \in \mathbf{O}$}
    \STATE $//$ shift the object depth
    \STATE $(x_m, y_m, z_m) = DECODE (O_m)$;
    \STATE $O_m \leftarrow ENCODE (x_m / r, y_m / r, z_m / r)$;
\ENDFOR
\ENSURE $\mathbf{I}, \mathbf{O}$
\end{algorithmic}
\label{algo:mstrain}
\end{algorithm}

\subsection{Head and Loss}

The detection head is a decoder-style process, where each object query $q_i$ predicts the new bounding box $\bb_i$ and its categorical label $\bc_i$ with two neural networks $\Theta^{reg}$ and $\Theta^{cls}$:
\begin{align}
	\bb_i = \Theta^{reg}(q_i^{new}), \\
	\bc_i = \Theta^{cls}(q_i^{new}).
\end{align}

Analogous to \cite{wang2022detr3d}, we adopt the focal loss \cite{lin2017focal} for classification and L1 loss \cite{ren2015faster} for 3D bounding box regression. Given on the loss computation between the ground-truth $y=(c, b)$ and the prediction $\hat{y} = (\hat{c}, \hat{b})$, the Hungarian algorithm is used for the optimal label assignment. To find a bipartite matching between these two sets, we search a permutation of $\hat{\sigma}$ with the lowest cost:
\begin{equation}
	\hat{\sigma} = \text{arg min} \sum_i^N \mathcal{L}_{match}(y_i, \hat{y}_i),
\end{equation}
where $\mathcal{L}_{match}(y_i, \hat{y}_i)$ is a pair-wise matching cost. And the final objective function is derived as:
\begin{equation}
	L(y, \hat{y}) = L_{cls}(c, \hat{\sigma}(\hat{c})) + L_{reg} (b, \hat{\sigma}(\hat{b})).
\end{equation}

\section{Experiments}

\subsection{Dataset}
The nuScenes dataset \cite{caesar2020nuscenes} is one of the most popular datasets for 3D object detection, consisting of 700 scenes for training, 150 scenes for validation, and 150 scenes for testing. For each scene, it includes 6 camera images (\textit{front, front\_left, front\_right, back\_left, back\_right, back}) to cover the whole viewpoint. Camera parameters including intrinsics and extrinsics are available, which establishes a one-to-one correspondence between each 3D point and the 2D image plane. There are 23 categories in total, among which only 10 classes are utilized to compute the final metrics.

\subsection{Implementation Details}
The model encoder consists of a ResNet \cite{he2016deep} backbone, a FPN neck, and a DETR-style detection head. We use ResNet50 with deformable convolutions \cite{dai2017deformable} adopted in the 3$^{rd}$ and 4$^{th}$ stage of the residual block by default. The FPN provides hierarchical image feature maps at a resolution of $1/8, 1/16, 1/32,$ and $1/64$ of the input image size. The detection head of Graph-DETR3D follows the design of DETR, consisting of 6 layers, where each layer is a combination of our dynamic graph-based feature aggregation module and a multi-head attention module. We use AdamW optimizer and the weight decay is 10$^{-4}$. The initial learning rate is set to $2\times 10^{-4}$ and decayed in the cosine annealing schedule. The model is trained for 12 epochs in total on 8 Tesla V100 GPUs with a batch size of 8.

\subsection{Main Results}
\subsubsection{Detection Results in Different Regions} 
We first implement DETR3D and Graph-DETR3D on nuScenes validation subset with 1$\times$ schedule. The final results evaluated based on different split regions are shown in Table \ref{tab:nus_val}. Our Graph-DETR3D greatly boosts its vanilla DETR3D by more than 1.9/1.5 \% on NDS score and mAP, respectively. Such improvements validate the effectiveness of the proposed method. When taking a closer look at the result, we find the performance on the overlapping regions gets the most improvement (+3.1 mAP, +3.3 NDS), which proves the superiority of the proposed Graph-DETR3D on leveraging multi-view information.

\begin{table}[!h]
	\begin{center}
		\setlength{\tabcolsep}{0.5mm}
			\begin{tabular}{c | c c | c c}
				\toprule
				\multirow{2}*{Region}  &  \multicolumn{2}{c|}{DETR3D} & \multicolumn{2}{c}{Graph-DETR3D}\\
				   & mAP (\%) & NDS (\%) & mAP (\%)  & NDS (\%) \\
				\midrule
				Overall & 33.6 & 41.4 & 35.1 \footnotesize{(\textcolor{cyan}{$+1.5$})} & 43.3 \footnotesize{(\textcolor{cyan}{$+1.9$})}\\
				Non-Overlap & 33.3 & 41.1 & 34.1 \footnotesize{(\textcolor{cyan}{$+0.8$})} & 42.3 \footnotesize{(\textcolor{cyan}{$+1.2$})}\\
				Overlap & 25.0 & 38.0 & 28.1 \footnotesize{(\textcolor{blue}{$+3.1$})} & 41.3 \footnotesize{(\textcolor{blue}{$+3.3$})}\\
				\bottomrule
			\end{tabular}
	\end{center}
	\caption{Comparison of DETR3D and Graph-DETR3D in different regions on nuScenes validation subset.}
	\label{tab:nus_val}
	\vspace{-2em}
\end{table}

\subsubsection{Comparison with State-of-the-Arts}

In addition to offline results, we also report the detection performance on nuScenes test leaderboard compared with various detection approaches. The results are shown in Table \ref{tab:nus_test}. When using the same backbone network, our Graph-DETR3D beats all other methods, including the recently developed DETR3D \cite{wang2022detr3d} and BEVDet \cite{huang2021bevdet}. Note that our method with a ResNet101 backbone has already transcended the LiDAR-based approach, PointPillar \cite{lang2019pointpillars}, which implies its great prospect in various applications. Additionally, empowered by the advanced V99 backbone \cite{lee2020centermask}, our best model achieves a suppressing 49.5 NDS score in the single scale testing protocol, achieving new state-of-the-art performance on this competitive benchmark with camera sensors only.


\subsection{Ablation Studies}
\label{sec:ablation}
 In this section, we provide ablative experiments to gain a deeper understanding of Graph-DETR3D. The experiments are conducted with a ResNet50-DCN backbone evaluated on nuScenes validation subset. Following the experimental setting in \cite{wang2021pointaugmenting}, 1/4 nuScenes training set is used for efficiency.
To understand how each module contributed to the final detection performance in Graph-DETR3D, we test each component independently and report its performance in Table \ref{tab:ablation}. The overall score starts from 35.8 NDS. When dynamic graph feature aggregation module is applied, the NDS is raised by 2.4 points, which validates the finding in Section \ref{sec:mstrain} that enhancing the detection ability on overlapping regions can greatly promote the overall performance. Then we add the depth-invariant multi-scale training strategy, which brings us a 0.9 enhancement as well as a 2.8 \% decline in mAOE. Finally, the NDS score achieves 39.3 when all components are applied, gaining a 3.5 \% absolute improvement and validating our approach.


\begin{table}[!h]
    \begin{center}
    \setlength{\tabcolsep}{2mm}
    \begin{tabular}{c|cc|ccc}
        \toprule
        No. & DGFA & DI MS-Train & NDS$\uparrow$ & mAP$\uparrow$ & mAOE$\downarrow$ \\
        \noalign{\smallskip}
        \hline
        \noalign{\smallskip}
        1& & & 0.358 & 0.291 & 0.527 \\
        2&$\checkmark$& & 0.382 & 0.311 & 0.489\\
        3& &$\checkmark$& 0.367 & 0.300 & 0.499\\
        4& $\checkmark$ &$\checkmark$& \textbf{0.393} & \textbf{0.319} & \textbf{0.460} \\
        \bottomrule
        \end{tabular}
    \vspace{0.2em}
    \caption{Effectiveness of each component in Graph-DETR3D. Results are reported on nuScenes validation subset.}
    \label{tab:ablation}
	\end{center}
	\vspace{-2em}
\end{table}

%
%

\begin{table*}[t!]
\begin{center}
\setlength{\tabcolsep}{1.0mm}
\begin{tabular}{l|c|c|c|c|cccccc}
\toprule
\noalign{\smallskip}
Method & Modality & Backbone & Epoch & NDS$\uparrow$ & mAP$\uparrow$ & mATE$\downarrow$ & mASE$\downarrow$ & mAOE$\downarrow$ & mAVE$\downarrow$ & mAAE$\downarrow$ \\

\noalign{\smallskip}
\hline
\noalign{\smallskip}
 PointPillar \cite{lang2019pointpillars} & L & - & n/a & 0.453 & 0.305 & - & - & - & - & -\\
 CenterPoint \cite{yin2021center} & L & - & n/a & 0.655 & 0.580 & - & - & - & - & -\\
 \hline
 MonoDIS \cite{simonelli2019disentangling} & C & ResNet-50 & n/a & 0.384 & 0.304 & 0.738 & 0.263 & 0.546 & 1.553 & 0.134 \\
 FCOS3D$\ddagger$ \cite{wang2021fcos3d} & C & ResNet-101 & 24 &0.428 &0.358 &0.690 &0.249 &0.452 &1.434 &\textbf{0.124}  \\
 PGD$\ddagger$ \cite{wang2022probabilistic}& C  & ResNet-101 & 24 &0.448 &0.386 &0.626 &0.245 &0.451 &1.509 &0.127 \\
 DD3D$\ast\ddagger$ \cite{park2021pseudo}& C  & V2-99 & 24 &0.477 &0.418  &0.572 &0.249 &\textbf{0.368} &1.014 &\textbf{0.124} \\
 DETR3D$\ast$ \cite{wang2022detr3d}& C  & V2-99 & 24 &0.479 &0.412  &0.641 &0.255 &0.394 &0.845 &0.133 \\
 BEVDet \cite{huang2021bevdet}& C  & Swin-S & 24 &0.463 &0.398 &0.556 &\textbf{0.239} &0.414 &1.010 &0.153  \\
 BEVDet$\ast$ \cite{huang2021bevdet}& C  & V2-99 & 24 &0.488 &0.424 &\textbf{0.524} &0.242 &0.373 &0.950 &0.148 \\
\hline
\textbf{(Ours)}& C & ResNet-101 & 24 & 0.472 & 0.418 & 0.668 & 0.250 & 0.440 & 0.876 & 0.139 \\
\textbf{(Ours)}$\ast$  & C & V2-99 & 24 &\textbf{0.495} &\textbf{0.425} & 0.621 & 0.251 & 0.386 & \textbf{0.790} &0.128 \\
\bottomrule

\end{tabular}
\end{center}

\caption{Comparison of recent works on the nuScenes testing set. $\ast$ are trained with external data and $\ddagger$ stands for test time augmentation. ``L'' and ``C'' indicate LiDAR and Camera, respectively.}
\label{tab:nus_test}
\end{table*}

\subsection{Discussion}

In this section, we delve into the Graph-DETR3D framework to study how the detection accuracy is achieved and gain a deeper understanding of the underlying mechanisms. For all experiments, we employ the same experimental settings in Section \ref{sec:ablation}.

\textbf{Investigating the Best Feature Aggregation Operation}. In this part, we compare the DGFA module with various strategies for multi-view features aggregation. Firstly, we choose the single-point projection, which is proposed in DETR3D \cite{wang2022detr3d}, as our baseline. Then, we further extends this paradigm into a multi-point sampling process by a constructing unified graph structure. Specifically, the 8 corners are decoded based on the predicted bounding box and the corresponding reference point to form a fixed subgraph. We also compare with a general form of deformable convolution, where the reference point is projected into the image plane and dynamically aggregate features with learnable offsets. The detailed results are listed in Table \ref{tab:abla_dgfa}. The single-point feature sampling baseline is 29.1 mAP. When adopting the multi-point feature aggregation, the improvement is limited. Though the object query can perceive more information through graph message propagation, the sampling features may not be suitable due to the inaccurate bounding boxes estimation. Compared to vanilla multi-point feature aggregation, deformable feature aggregation paradigm enables the network to automatically decide the sampling position at the 2D image plane, which gets prospective advancement. However, it still suffers from the dilemma that it fails to aggregate multi-view features when the reference point is projected into a single camera-view. Our DGFA module addresses such issue by constructing a dynamic graph in the 3D space and simultaneously refines the graph topologies and object queries, which achieves the best performance (31.9 mAP).

\begin{table}[!h]
	\begin{tabular}{c|ccc}
        \toprule
        Method & NDS$\uparrow$ & mAP$\uparrow$ & mAOE$\downarrow$\\
        \noalign{\smallskip}
        \hline
        \noalign{\smallskip}
        single point \cite{wang2022detr3d} & 0.358 & 0.291 & 0.527\\
        fixed multi-point & 0.375 & 0.302 & 0.483\\
        deformable attention \cite{zhu2020deformable} & 0.388 & 0.311 & 0.468\\
        DGFA (Ours) & \textbf{0.393} & \textbf{0.319} & \textbf{0.460} \\
        \bottomrule
    \end{tabular}
    \vspace{0.4em}
    \caption{Detection performance of different approaches for imagery feature aggregation on nuScenes validation subset.}
    \label{tab:abla_dgfa}
\end{table}

\textbf{Does Multi-Scale Training Really Improve the Detector?} Multi-Scale training is a common practice in object detection. Empirically, the enhancements brought by multi-scale training can be grouped into two aspects: (1) feature extraction ability of the image backbone, and (2) object localization ability of the detection head. In Section \ref{sec:mstrain}, we have analyzed the deterioration effect on 3D detectors caused by the ambiguous supervision signals. Therefore, to validate if depth-invariant multi-scale training algorithm truly overcomes this issue is nontrivial, as such improvement may come from the enhancement of feature extraction only. We first implement a vanilla multi-scale training strategy, where only image size and camera intrinsics are changed. Considering that the classification task won't lead to the ambiguous localization issue, we devise a disentangled multi-scale training implementation where only the classification branch is supervised if the input image size is changed. Otherwise, both classification and regression loss are propagated. The final results are reported in Table \ref{tab:abla_mstrain}. The model performance gets a 2.7 mAP performance drop when vanilla multi-scale training is adopted, indicating that the enhancement on feature extraction can not mitigate the deterioration effect on the detection head. When replacing it with the advanced multi-scale training strategy, the ambiguous issue is greatly alleviated, which brings a 0.2 mAP improvement. However, it misses the chance to fully boost the localization ability of the detection head under multi-scale conditions. Conversely, the depth-invariant multi-scale training strategy tackles this issue directly by simultaneously rescaling the image size and the depth supervision and gains the most improvement. 

\begin{table}[!h]
	
	\begin{tabular}{c|ccc}
        \toprule
        Method & NDS$\uparrow$ & mAP$\uparrow$ & mAOE$\downarrow$\\
        \noalign{\smallskip}
        \hline
        \noalign{\smallskip}
        baseline & 0.358 & 0.291 & 0.527\\
        vanilla MS-Train & 0.331 & 0.262 & 0.539 \\
        disentangled MS-Train  & 0.360 & 0.294 & 0.522\\
        DI MS-Train & \textbf{0.367} & \textbf{0.300} & \textbf{0.499}\\
        \bottomrule
    \end{tabular}
    \vspace{0.5em}
    \caption{Detection performance with different multi-scale training strategies on nuScenes validation subset. }
    \label{tab:abla_mstrain}
    \vspace{-1em}
\end{table}

\section{Conclusion}
In this work, we develop a new paradigm for multi-view 3D detection, Graph-DETR3D. It fully leverages the information from surrounding cameras with a novel dynamic graph feature aggregation module, which significantly improves the detection performance. Besides, we explore a general data augmentation form in vision-based 3D detection, named depth-invariant multi-scale training. 
The experiments demonstrate that Graph-DETR3D achieves state-of-the-art results on the challenging nuScenes benchmark, and validate the effectiveness and efficiency of our model. 
We hope Graph-DETR3D can serve as a solid baseline and provide a new perspective for multi-camera perception tasks.

{\small
\bibliographystyle{ieee_fullname}
\bibliography{graph-detr3d}
}

\end{document}